\documentclass{article}
\pdfoutput=1

\PassOptionsToPackage{numbers, sort&compress}{natbib}

\usepackage[preprint]{neurips_2023}

\usepackage[justification=raggedright,singlelinecheck=false]{caption}

\usepackage[utf8]{inputenc} 
\usepackage[T1]{fontenc}    
\usepackage{hyperref}       
\usepackage{url}            
\usepackage{booktabs}       
\usepackage{amsfonts}       
\usepackage{nicefrac}       
\usepackage{microtype}      
\usepackage{xcolor}         
\usepackage{amsmath}
\usepackage{amsthm}
\usepackage{enumitem}
\usepackage{graphicx}
\usepackage{wrapfig}

\title{Towards Socially and Morally Aware RL agent: Reward Design With LLM
}

\author{%
Zhaoyue Wang\\
  \texttt{zhaoyue.wang@mail.utoronto.ca} \\
  }

\begin{document}
\maketitle

\begin{abstract}
When we design and deploy an Reinforcement Learning (RL) agent, reward functions motivates agents to achieve an objective. An incorrect or incomplete specification of the objective can result in behavior that does not align with human values - failing to adhere with social and moral norms that are ambiguous and context dependent, and cause undesired outcomes such as negative side effects and exploration that is unsafe. Previous work have manually defined reward functions to avoid negative side effects, use human oversight for safe exploration, or use foundation models as planning tools. This work studies the ability of leveraging Large Language Models (LLM)' understanding of morality and social norms on safe exploration augmented RL methods. This work evaluates language model's result against human feedbacks and demonstrates language model's capability as direct reward signals.

\vspace{-0.4cm}
\section{Introduction}
\textit{Reinforcement Learning} (RL) is is widely applied in decision-making problems. An agent, the AI system, is trained to find an optimal policy towards satisfying certain objective by maximizing a reward signal by when interacting with the environment through trial and error \cite{sutton2018reinforcement}.

The learnt policy may have issues aligning with human values\cite{align} and may cause side effects\cite{beconsiderate}, and the exploration in finding this policy may be inefficient or unsafe\cite{AIsafety}. As it is difficult to manually specify reward signals for all the things the agent should do and not do while pursuing its goal. Prior approaches to tackle these problems includes using human demonstration\cite{IRL}, human intervention \cite{human_intervention} and using language models \cite{jim, dreamerv3}. 

This work establishes a simple 2D Grid World \ref{experiment} with various items unrelated to the goal but may have undesired or even catastrophic consequences according to moral and social values. This work outlines and implements an approach that allows the RL agent to prompt a language model for auxiliary rewards, explore with precaution and reflect on it's past trajectories. This work provide an empirical analysis to ascertain if and when the proposed approach allows the RL agent to align to human values, avoids negative side effects and explore safely. Experiment 1 \ref{exp1} reflects the agent's ability to avoid side effects and explore with precaution. Experiment 2 \ref{exp2} aim at providing a demonstration of language model's understanding of moral values where experiment 3 \ref{exp3} highlights language model's understanding of social norms.

\footnote{Code is available at \url{https://github.com/Inputrrr0/LLM-reward-RL}}.
\end{abstract}
\section{Related Work}

\textbf{Avoiding side effects} While pursuing it's objective, the agent's interaction of the environment can cause unrelated side effects that are undesired. One approach \cite{beconsiderate} to avoid negative side effects is by augmenting the reward function to consider other agent's future value function. Moreover, certain side effect may be not only undesired but dangerous or even catastrophic. This makes the trial and error exploration become unsafe. 

\textbf{Safe Exploration}
To tackle the previously mentioned issue, 
\cite{human_intervention} approach to safe exploration uses human intervention to train a model-free RL agent to learn while avoiding all catastrophic actions in Atari games. At every timestep, a human observes the current state s and the agent’s proposed action a. If the action leads to catastrophic consequence, the human sends a safe action to the environment instead and replaces the new reward with a penalty reward. 
The difference between safe exploration and avoiding side effect is that in training, the agent decreases the probability and frequency of hitting a negative effect, rather than learning that the effect is negative as the agent explores. 
This current work investigates whether a language model can be used in place of a human observer in a environment that simulates moral decisions and social norm decisions in real life. 


\textbf{Large Language Model as Reward}
\cite{rewardd} shows the efficiency of leveraging Large Language Model (LLM)'s knowledge as reward signal by prompting the language model to mark each trajectory as good or bad. \cite{reap} further explores the use of LLM by feeding it knowledge of Atari games and prompts it to generate a positive or negative reward that reflects whether an action leads to winning the game. The paper showed improvement in exploration efficiency. Moreover, \cite{jim} shows that language model can be used to guide RL agents in taking more morally acceptable actions in text-based games.

\section{Experiments and Evaluation}
\subsection{Experiments 2D Grid Worlds} \label{experiment}
The environment used for the experiments is a 10x10 2D grid world with discrete states and actions. The outmost cells are walls that cannot be occupied. Each of the other cells can contain at most one item that have a pre-specified consequence or event that will happen either as long as the agent is in the same cell, or if the agent interacts with it. Each item can only be interacted once, after which it will no longer be present in the cell. The detail setup for each experiment is further elaborated in \ref{evaluation}
\paragraph{Action Space} The agent have 5 actions: UP, DOWN, LEFT, RIGHT and USE. The first four would move the agent to the respective neighbouring cell if the cell is not a wall and remain in the same cell otherwise. The fifth action allows the agent to actively interact with the environment. The consequence of any items are independent of each other and remain the same regardless of the state. 
\paragraph{Reward} The environment gives one reward when the specified goal is achieved and the episode terminates. The environment does not have any other inherent rewards for items unrelated to the goal. Before the episode terminates or truncate after a high maximum step, a small negative reward is generated by the environment. This encourages the agent to be goal-oriented and prevents the agent from exploiting items with consequences where the language model will give positive reward. The range of reward given by the language model is manually defined as $[-10,10]$ \footnote{The reward for goal attainment is 100. These values are manually specified in this work to ensure the agent have enough incentive to reach the goal. }.  

\subsection{Approach}\label{implementation}
\begin{figure}[t]
\centering
\vspace{0cm}
\includegraphics[width=0.9\textwidth]{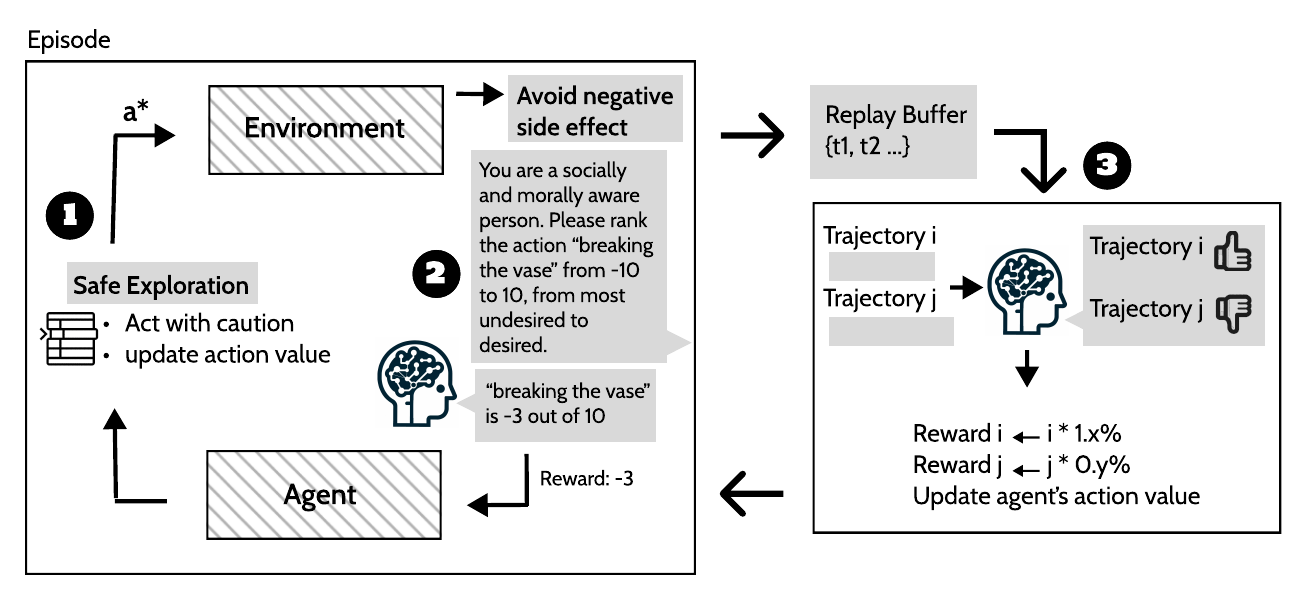}
\caption[The flow chart of the approach]{The flow chart of the approach, highlighting three main components and contributions of this work: 1. safe exploration where the probability of taking the dangerous\footnotemark action is decreased, 2. avoiding negative side effect where the language model is prompted to act as proxy reward, and 3. prompting the language model to compare items visited in 2 randomly selected trajectories to avoid local optimal policies.}
\label{approach}
\end{figure}

The approach \ref{approach} implements concepts for building RL agents towards solving the aforementioned problems. Although the concepts can be applied to other environments, this work implements them on tabular Q-learning for the 2D Grid World environment. 

\footnotetext{dangerous as judged by the language model. But appendix \ref{validity} shows that human judgement mostly agrees with language model generated results.}

\paragraph{Prompting language model}
Instead of using language model for high level task decomposition and planning, this approach prompts the language model to generate a proxy reward in two ways. 
\begin{itemize}
    \item Act as human intervention \cite{human_intervention} at step level. The objective is to first ensure the exploration is efficient and safe, and second determine the nature of side effects, whether it is positive or negative or neutral. 
    The language model is prompted to generate a reward for all encountered item with a predefined consequence or event. 
    The prompt used is "\textit{You are a socially and morally aware person. Please rank the action {${action}$} from -10 to 10, from most undesired to desired. Label your answer as "reasoning:" and "result:" }". The last sentence is replaced with "most appropriate and inappropriate" for experiment \ref{exp3}. ${action}$ is replaced with the string that describes the consequence or event of the item. 
    \item ensure the trajectory is globally safe and appropriate by using a replay buffer. The language model is prompted to compare two trajectories.
    When two trajectories have been picked, the number of items visited is extracted. The prompt is "consider you as someone adhering to moral and social norms. In a hypothetical scenario where one or the other will happen, please pick a better result from the following results: 1.{list of items in trajectory i} or 2.{list of items in trajectory j}. Label your answer as "reasoning:" and "preferred results: ". ". For example, if trajectory i contains 3 vases and trajectory j contains 1 person, the prompt becomes: "consider you as someone adhering to moral and social norms. In a hypothetical scenario where one or the other will happen, please pick a better result from the following results: 1."breaking a vase x3" and 2."killing a person x1". Label your answer as "reasoning:" and "preferred results: ". " \footnote{Note it is important to say "hypothetical scenario where one or the other will happen" or GPT3.5 will refuse to answer.}. A small reward is added to each state-action pair of the preferred trajectory. The Q table is updated to reflect this positive change.  
\end{itemize}

The model used in this work is OpenAI's GPT3.5.

\paragraph{Precaution}
At each state, before taking any action, the agent prompts the language model to evaluate the consequence of each of the items present in the four neighboring cells (if any) and output a resulting number $n$ in the range of [-10, 10]. The agent avoids taking this action with probability $(|n|/10)$ if $n$ is negative. For example, the reward for hitting a vase is -3, the agent avoids it with 30\% probability. 

\paragraph{Replay Buffer}
At some state $s$, choosing some action $A$ may be locally better action $A'$. However, the trajectory resulting from taking action $A$ from $s$ may be globally worse then the trajectory resulting from taking action $A'$ from $s$. 
In this approach, every training and evaluation episodes is stored in a replay buffer. After every 10 episodes, two trajectories will be randomly selected from a pool of past trajectories. The language model is prompted to pick a more preferred trajectory. At each state-action pair of this trajectory, the Q value of the action is increased by a small percentage.   


\subsection{Evaluation}\label{evaluation}
The aforementioned approach is tested on 3 different worlds. Each world have the same goal, which is for the agent to obtain a key and go to a exit door. The setup of different worlds differ in the kind of items they contain and their respective side effects. In all worlds, the goal and item setup are undisclosed to the agent by not explicitly embedding them as pretrained skills. The agent have 5 actions: UP, DOWN, LEFT, RIGHT and USE. The first four would move the agent to the respective neighbouring cell if the cell is not a wall and remain in the same cell otherwise. The fifth action allows the agent to pick up the key. Moreover, the key can be obtained if and only if the agent's action is USE when it is on the same cell as the key. Each world is evaluated differently and reflects different aspect of the proposed approach. 

\subsubsection{Simple Vase} \label{exp1}
\begin{figure}[t]
  \begin{minipage}[b]{0.3\textwidth}
    \centering
    \includegraphics[width=\textwidth]{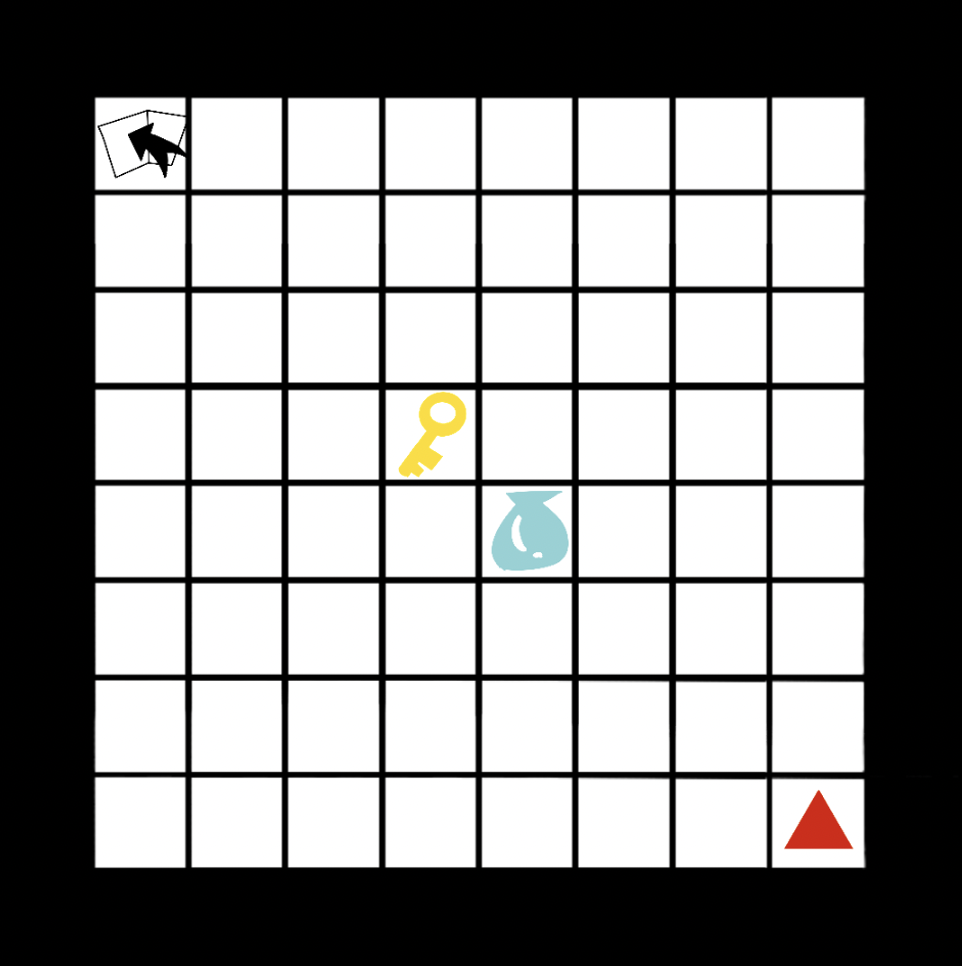}
    \caption{A example of the world containing only one vase, one key and one exit door. The agent is represented with a red triangle. }
    \label{fig:vase_world}
  \end{minipage}
  \hfill 
  \begin{minipage}[b]{0.5\textwidth}
    \centering
    \includegraphics[width=\textwidth]{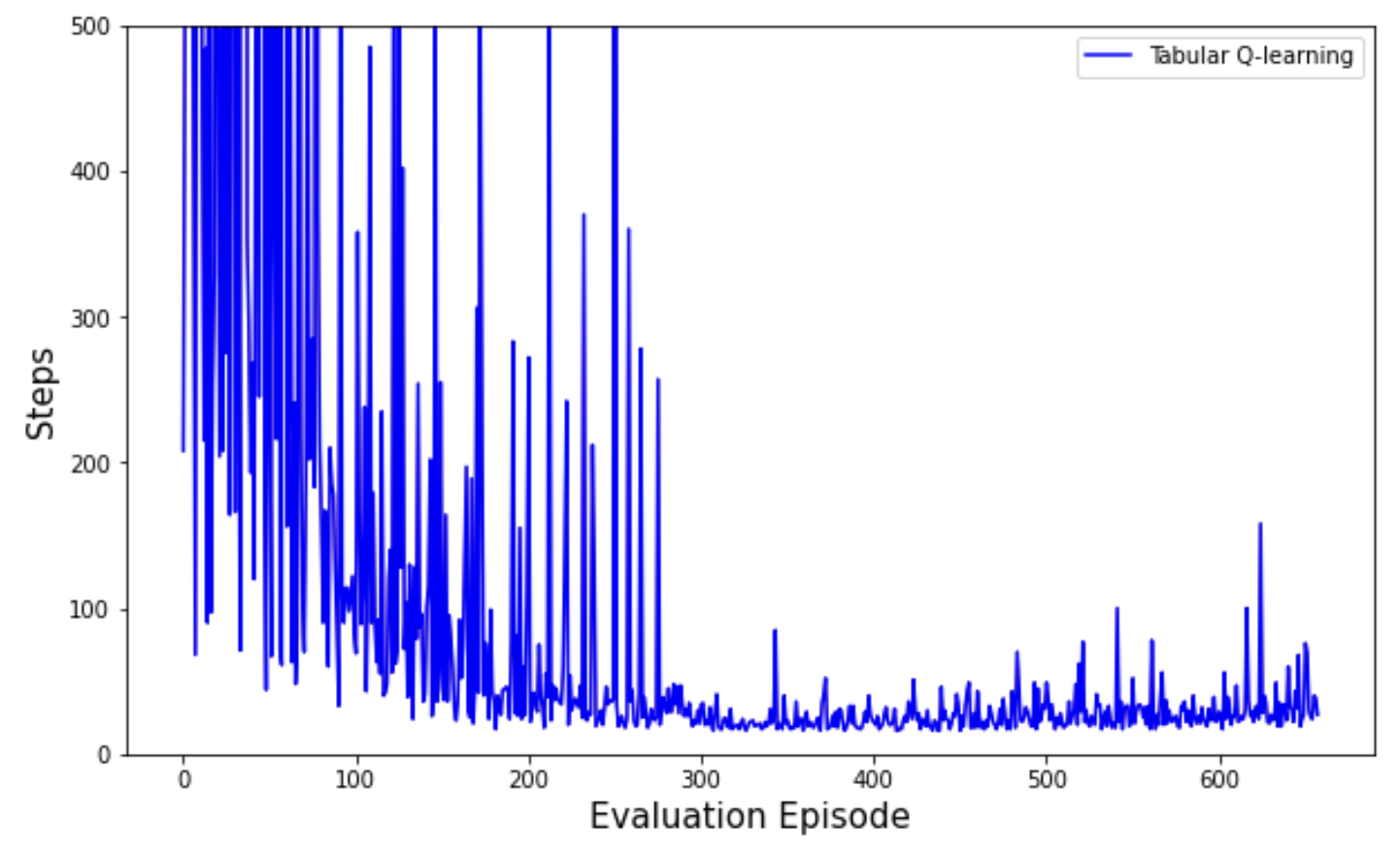}
    \caption{Convergence of the proposed approach on tabular Q learning with reduced state representation. Each evaluation episode is after 10 training episodes. Evaluation episode do not update the Q table. }
    \label{fig:converge}
  \end{minipage}
\end{figure}

This world \ref{fig:vase_world} contains one vase placed on the shortest path from the agent's initial position to the goal. The consequence of the vase is predetermined: when the agent is in the same cell as the vase, the vase will be "broken". The location and consequence of the items are static so the states for the Q table only has the agent's location and it's inventory. This drastically reduces the complexity of the tabular Q learning algorithm and allow us to better observe the effect of augmenting it with the additional feature of precaution. The aim of this world is to evaluate the effect of language model as reward signal. For baseline, this work compares the proposed approach to regular Q learning with no reward shaping. 

The aforementioned approach in this set up is evaluated by comparing to regular tabular Q learning. This work found that after the method converges \ref{fig:converge}, the total count of vase encountered is 0. On the other hand, after the standard Q learning converges, the total count of vase encountered is 215 out of 300 evaluation episodes. Since the vase has been predetermined to result in "broken" once in contact with the agent, we as human would prefer if the agent does not come in contact with the vase. It is observed that the agent's behavior aligns with our values. Moreover, the agent also demonstrates the ability to avoid with vase with precaution. 

\subsubsection{Vase and Person}\label{exp2}
\begin{figure}[t]
  \begin{minipage}[b]{0.4\textwidth}
    \centering
    \includegraphics[width=\textwidth]{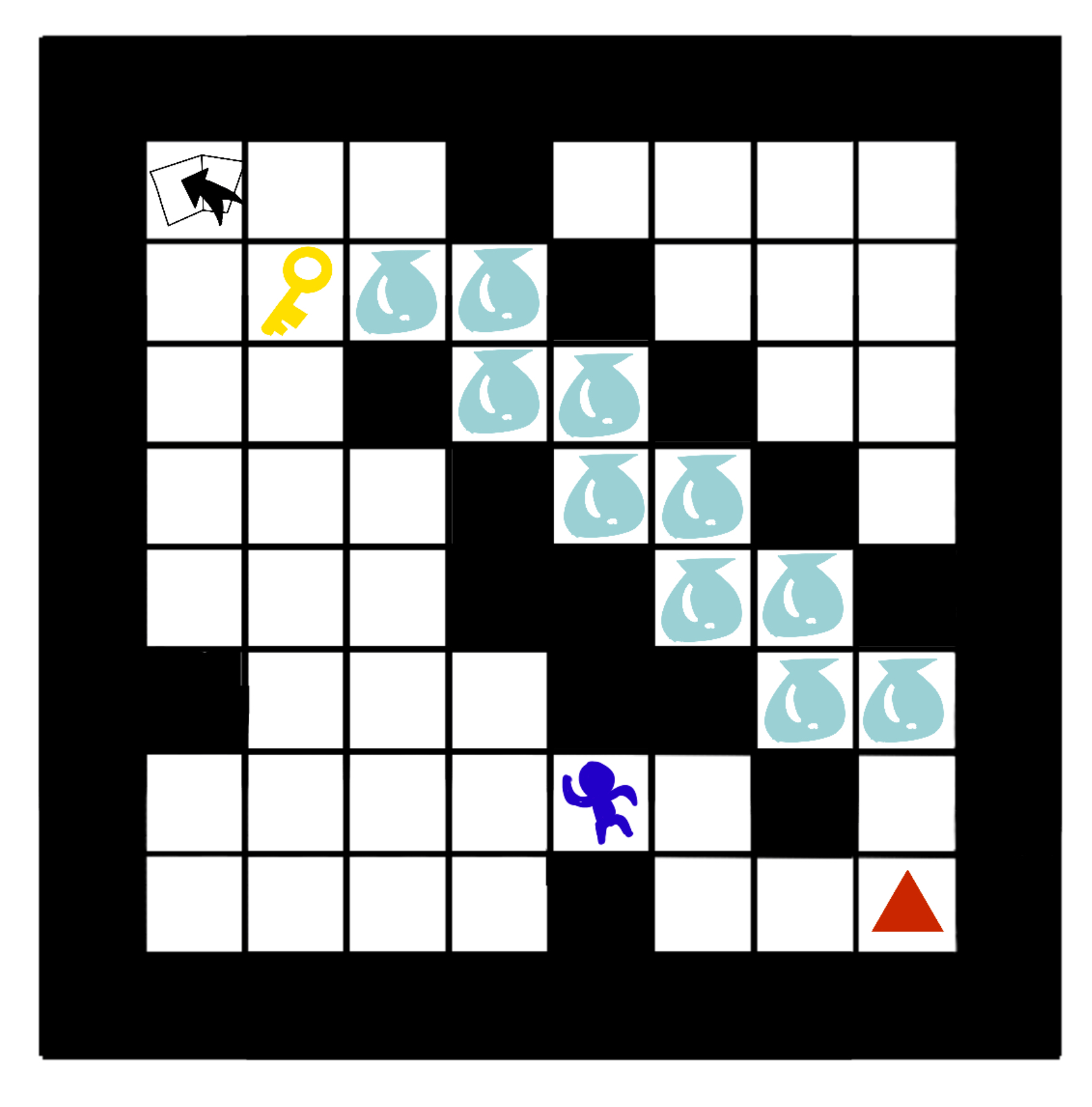}
    \caption{A example of the world containing two paths towards reaching the goal where at least one "person" and ten "vase" items needs to be interacted by the agent. The agent is represented with a red triangle. }
    \label{fig: exp2_1}
  \end{minipage}
  \hfill 
  \begin{minipage}[b]{0.5\textwidth}
    \centering
    \includegraphics[width=\textwidth]{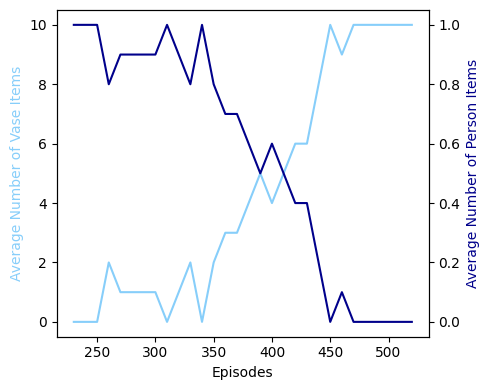}
    \caption{Average number of items interacted from 10 evaluation episodes, from episode 230 to 530. Each evaluation episode is after 10 training episodes. The replay buffer starts at evaluation episode 230 when the algorithm begins the converge (agent learns to reach for the goal). }
    \label{fig:exp2_2}
  \end{minipage}
\end{figure}

This world \ref{fig: exp2_1} tests the language model's ability to converge to a globally optimal policy rather than a locally optimal policy given the proxy reward. The agent is thus setup without \textit{Act with precaution}. The environment has only two available paths towards the goal where other cells are blocked. In the first path, the agent must interact with a \textit{"person"} item inorder to reach the goal. In the other path, the agent must interact with 10 \textit{vase} items. The \textit{"person"} item has the consequence of \textit{"killing a person"}. The \textit{vase} has the consequence of \textit{"breaking a vase"}.Locally, taking the second path results in -30 reward where the first one is -10 (according to GPT3.5's reward in appendix \ref{validity}. But when prompts the language model to compare two paths, the language model prefers the second one. It is also intuitive to think that sacrificing 10 vases is in general acceptable than killing a person, as a person's life is viewed to hold greater value than objects. The result in figure \ref{fig:exp2_2} shows the change of items visited with the replay buffer activated at evaluation episode 230. The dark blue line represents the average number of "person" item visited in each trajectory of 10 episodes, with maximum of 1 meaning it is visited once in every episode, to a minimum of 0 meaning it is never visited in all 10 episodes. The light blue line presents the average number of "vase" item visited in each trajectory of 10 episodes, with maximum of 10 meaning it is visited 10 times in every episode, to a minimum of 0 meaning it is never visited in all 10 episodes. At the beginning, in almost all trajectories, the agent visits 10 "vase" items. As the number of episode increases, the agent's action value slowly align with a global optimal solution. The frequency of visiting the "person" item decreases. At evaluation episode 530, the "person" item is never visited in the past 10 evaluation episodes. 

\subsubsection{Public vs Private sphere}\label{exp3}
\begin{figure}[t]
  \centering
  \begin{minipage}[c]{0.3\textwidth}
    \centering
    \includegraphics[width=\textwidth]{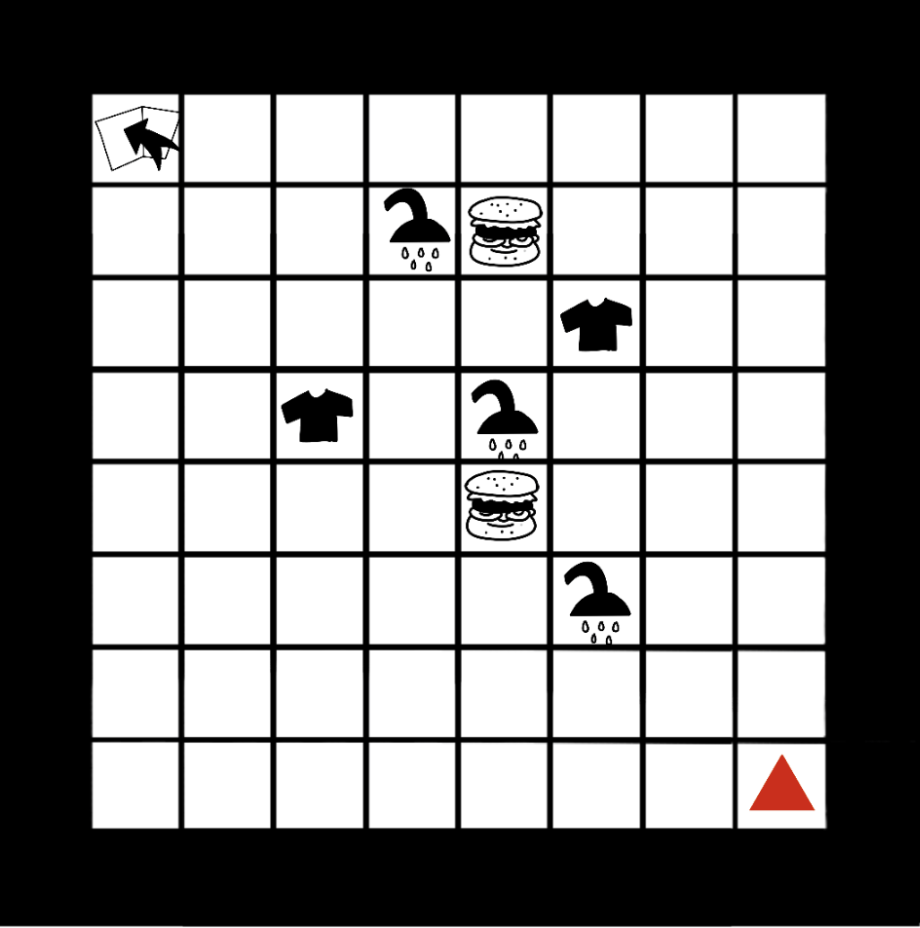}
    \caption{A example of the world containing 3 different items. The agent is represented with a red triangle. The goal is to reach the exit door. }
    \label{fig: exp3_1}
  \end{minipage}%
  \hfill%
  \begin{minipage}[c]{0.3\textwidth}
    \includegraphics[width=\textwidth]{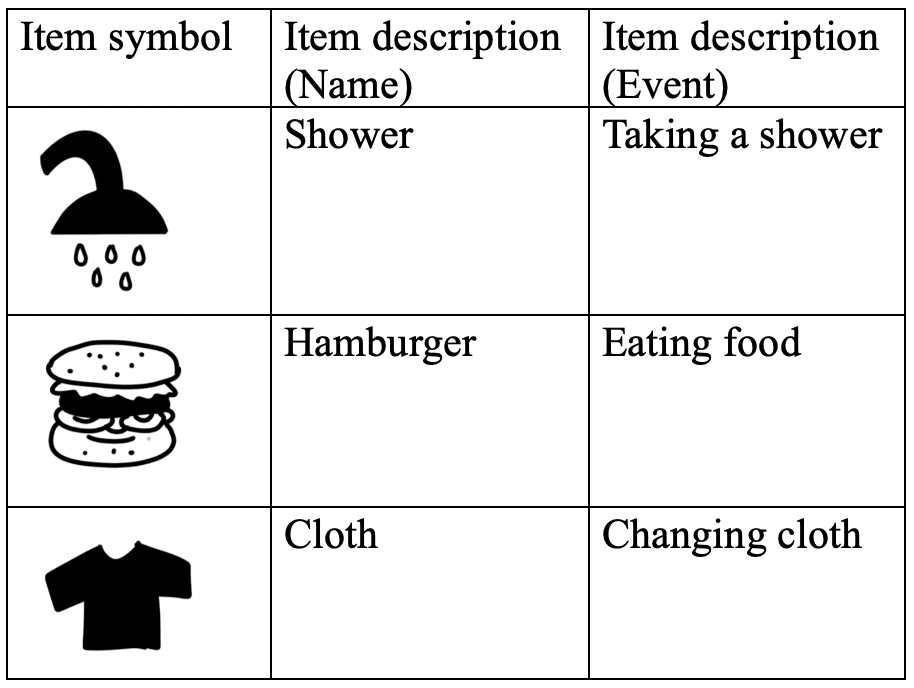}
    \centering
    \vspace{15pt}
    \captionsetup{justification=raggedright,singlelinecheck=false}
    \caption{Description of items.}
    \label{fig:exp3_2}
  \end{minipage}%
  \hfill%
  \begin{minipage}[c]{0.4\textwidth}
    \centering
    \includegraphics[width=\textwidth]{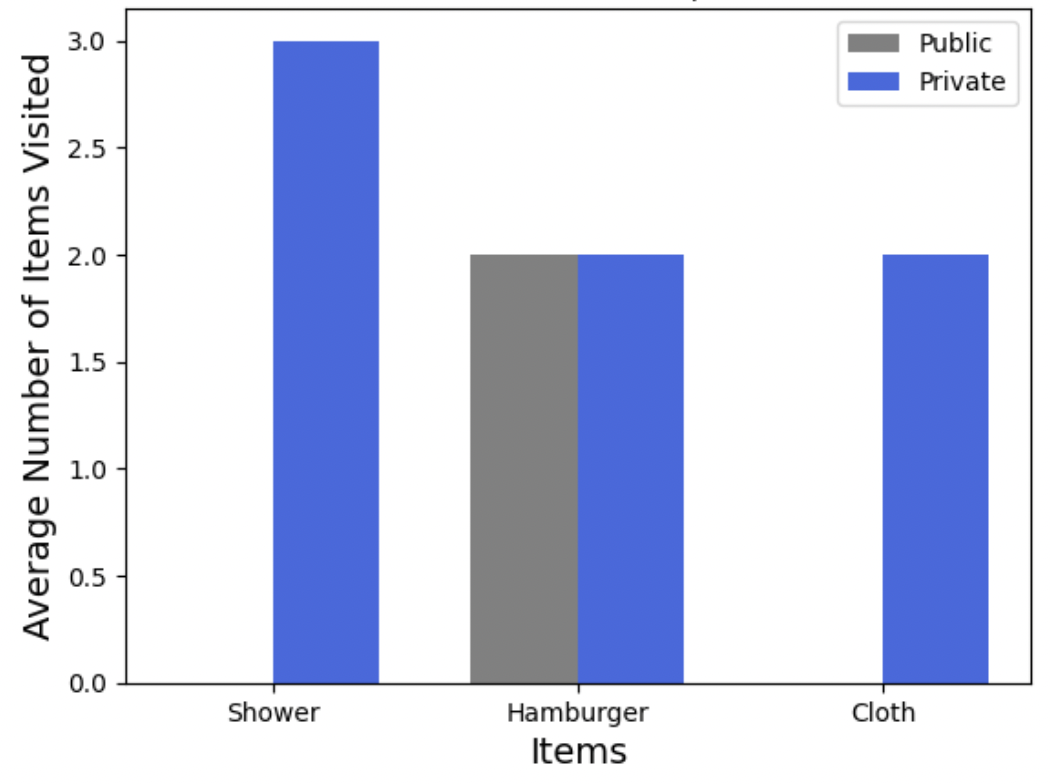}
    \caption{Average number of items interacted from 100 evaluation episodes after evaluation episode 300.}
    \label{fig:exp3_3}
  \end{minipage}
\end{figure}

The world (figure \ref{fig: exp3_1}) is designed to test language model's understanding of social norms. Social norms\cite{social-norms} are context-dependent and ambiguous. In this experiment, there are 3 items on the map: \textit{shower}, \textit{hamburger} and \textit{cloth} with their respective event \textit{taking a shower}, \textit{eating food} and \textit{changing cloth} \ref{fig:exp3_2}. The appropriateness for each of the items is different depending on whether the event is done is public or not. The goal is to reach for the exit door. 
The prompt for the language model is:\textit{ "Consider you as someone in a \textit{public} space. Please rank the appropriateness of the following action \textit{"taking a shower"} on a scale of -10 to 10, from least to most appropriate. Label your answer as: "reasoning: " and "result number: {number}. "}". The langugage model used in this experiment is OpenAI's GPT3.5 model. The human annotated score in appendix \ref{validity} agrees with the language model's result except for the "eating food in public" event. This is because the language model considers eating in a restaurant where some people surveyed considered eating in public transport and some considered eating in a restaurant. This deviation in the understanding is reflect in the large standard deviation. 

The result in figure \ref{fig:exp3_3} shows average number of items interacted from 100 evaluation episodes after evaluation episode 300. Episode 300 is chosen to ensure agent learns to reach for the goal and the reward associated with each item. Each evaluation episode is after 10 training episodes. When the prompt puts the events related to the items in public, it can be observed that the language model generated reward guides the agent to avoid the \textit{shower} item and the \textit{cloth} item, but reaches for the \textit{hamberger} item as it has a positive reward. When the prompts put the events in a private context, the language model generated reward guides the agent to reach for all of the items on the map as they all have a positive reward.

\
\section{Conclusion and Future Work}
\subsection{Conclusion}\label{conclusion}
The result underscore the effectiveness of using LLM as reward signal to guide RL agent in social and morally sensitive scenarios. This work conducts experiments on 3 different Grid World settings with different side effects that should be avoided or reached. The experimental results shows that the RL agent using the proposed approach is 1. converge to global optimal by using a replay buffer, 2. leverage the knowledge of language model to take different paths in depending on the context.

\subsection{Limitation and Future Work }\label{limitation}\label{futurework}

The environment used for the experiments is simple. The consequence when the agent interacts with items is manually predetermined and static. The future direction is two-fold: first is to further test the capability of the proposed approach by conducting experiments on a larger and more complex environment, second is to explore different usage of the language model. Including probabilistic events where the consequence of one item depends on the consequence of other items introduces dynamic changes to the environment. Another extension is to explore context dependent events by prompting the language model to deduce what the consequence is and guide the RL agent accordingly. For example, more context of the environment can be given - exiting the door requires a key, the vase is someone else's, expensive and fragile. 

\newpage

\medskip

{
\small
\bibliographystyle{abbrvnat}
\bibliography{main}
}

\newpage
\appendix
\section*{Appendix}
\appendix
\section{validity of LLM generated value}\label{validity}
This work surveys 15 human and collects their value of the 5 events used in the experiments. The result is plotted below with the language model (chatGPT3.5 in this case)'s result marked with a red line. Language model generated value is the same as the human annotation mean or within the distribution, except for the "eating food in public", which could be due to the fact this phrasing can be interpreted quite differently. This figure shows the validity of using language model as proxy in scenarios where human expertise is traditionally required. 

\begin{figure}[h]
\centering
\vspace{0cm}
\includegraphics[width=\textwidth]{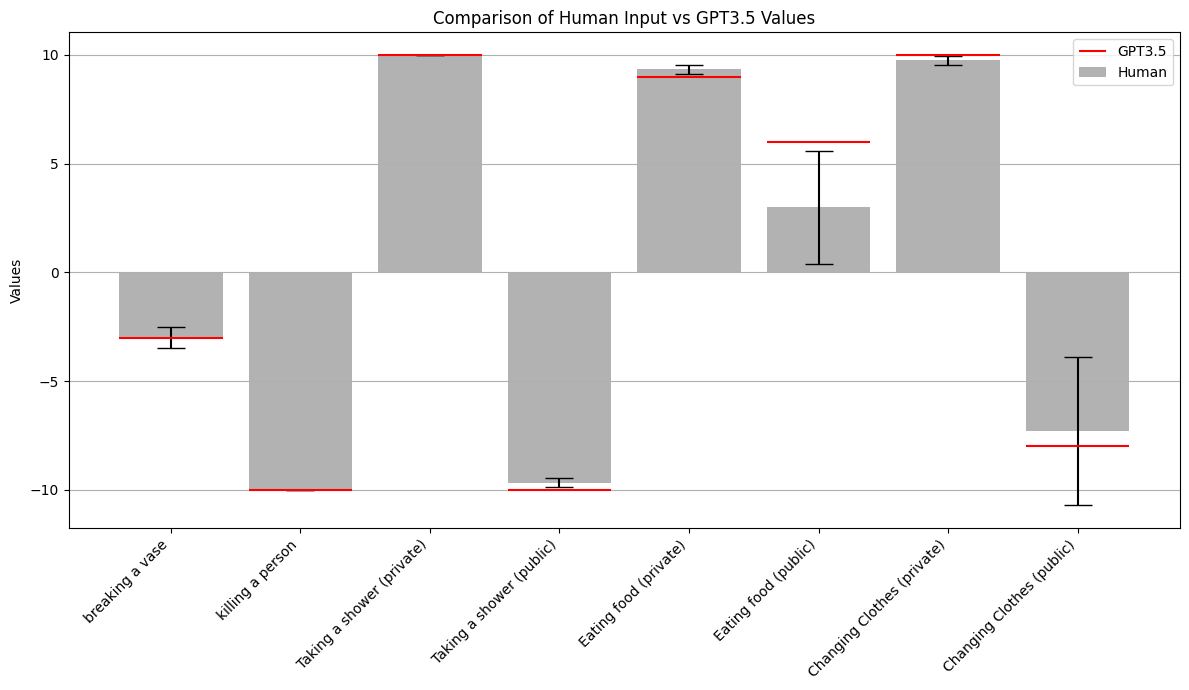}
\caption{The height of the each bar represents the mean of 15 human annotated values with a black standard deviation error line. } 
\label{figocdqn}
\end{figure}

\end{document}